\documentclass[conference]{IEEEtran}
\IEEEoverridecommandlockouts
\usepackage{mathptmx}

\usepackage[T1]{fontenc}
\usepackage[latin9]{inputenc}
\usepackage[letterpaper]{geometry}
\usepackage{graphicx}
\usepackage[dvipsnames]{xcolor}
\usepackage{textcomp}
\usepackage{soul}
\usepackage{placeins}
\usepackage{amsmath}

\usepackage[ruled,vlined]{algorithm2e}

\usepackage{graphicx}
\usepackage{subfigure}

\makeatletter
\usepackage{multirow}

\makeatother

\author{\author{
\IEEEauthorblockN{Kevin Brink, Ryan Sherrill, Jamie Godwin} 
\IEEEauthorblockA{
	Air Force Research Laboratory\\
	Munitions Directorate \\
    Eglin AFB, FL, 32542 \\
	Email: kevin.brink@us.af.mil
}

\and

\IEEEauthorblockN{Jincheng Zhang, Andrew Willis} 
\IEEEauthorblockA{
	Department of Electrical and Computer Engineering\\
	University of North Carolina at Charlotte\\ 
	Charlotte, NC 28223-0001\\
	Email: arwillis@uncc.edu
}}}
\begin{document}
\title{Maplets: An Efficient Approach for Cooperative SLAM Map Building Under Communication and Computation Constraints
\thanks{DISTRIBUTION A: Approved for public release.}
}
\maketitle

\begin{abstract}
This article introduces an approach to facilitate cooperative exploration
and mapping of large-scale, near-ground, underground, or indoor spaces
via a novel integration framework for locally-dense agent map data.
The effort targets limited Size, Weight, and Power (SWaP) agents with an emphasis on limiting required communications and redundant processing. 
The approach uses a unique organization of batch optimization engines
to enable a highly efficient two-tier optimization structure. Tier
I consists of agents which create and potentially share local maplets
(local maps, limited in size) which are generated using Simultaneous Localization and Mapping (SLAM) map-building software and then marginalized to a more compact parameterization. Maplets are generated in an overlapping manner and used to estimate the transform and uncertainty between those overlapping maplets, providing an accurate and compact odometry or delta-pose representation between maplet's local frames. The delta poses can be shared between agents, and in cases where maplets have salient features (for loop closures), the compact representation of the maplet can also be shared.

The second optimization tier consists of a global optimizer that seeks to optimize those maplet-to-maplet transformations, including any loop closures identified. This can provide an accurate global ``skeleton'' of the traversed space without operating on the high
density point cloud.   This compact version of the map data allows for scalable, cooperative exploration with limited communication requirements where most of the individual maplets, or low fidelity renderings, are only shared if desired. 
\end{abstract}

\begin{IEEEkeywords}
3D SLAM, UAV SLAM, quadcopter, large scale SLAM, graph SLAM 
\end{IEEEkeywords}

\section{Introduction}

This paper provides an approach for scalable, collaborative navigation, exploration, and mapping of an indoor or underground space. In many cases this navigation or mapping  work is done by a single agent. This imposes a computational burden on the agent to conduct all its processing on-board and  observations from a single reference often become brittle over time as odometry estimation errors begin to compound \cite{Indelman2012, Olson2006}. Multi-agent approaches to cooperatively navigate a space can outperform single agent approaches by taking advantage of additional information provided by various observations \cite{Sahawneh2017}. Collaborative approaches rely on coordinating in teams to rectify pose estimates which improves localization accuracy and provides opportunities for loop closures \cite{Melnyk2012, Kurazume1994}. In most cases, this requires a substantial amount of data to be shared so techniques to filter information are employed \cite{Sharma2008}. The reliance on a communication mechanism also demands a reliable solution to overcome communication issues between agents such as delays or dropout \cite{Strader2016}. 

Most of the recent collaborative approaches are based on extensions of graph-SLAM techniques such as \cite{Carlone2014,6696650,6631400}.  These references show that careful consideration
must be given to size, structure, and content of graph nodes for graph-SLAM
optimization.  Naive implementations will quickly exhaust either host
computing or memory resources \cite{Jimenez2018,Zhang2018} which can become detrimental in the cooperative mapping scenario. 

In this
work, we formulate a tiered optimization strategy to address some of those challenges and afford extreme-scale
navigation and mapping contexts while operating with limited communication and computational
resources. This effort is an adaptation of the cooperative navigation concept applied in \cite{Ellingson2020_Navigation} where IMU and camera data are fused into a high rate filter for a relative navigation framework that occasionally resets the filter reference frame to a new ``keyframe" \cite{Koch2020}. In this case, the high rate filter produces a marginalized delta-pose estimate, which represents the change in position and orientation, and covariance, the associated uncertainty between one keyframe and the subsequent keyframe. In this work, the corollary delta-pose measurement will be generated between one maplet and the next. 

When done well, the marginalized delta-pose measurements, effectively a low rate odometry measurement, and uncertainties provide an accurate representation of how the vehicle has moved over time \cite{Ellingson2020_Navigation}. They are also extremely compact compared to the large amounts of IMU, camera, or other data inputs.  They can also be easily placed in a batch optimization along with any additional relevant measurements to estimate a global position of the vehicle. By breaking up the nominal filter with respect to these keyframes, it accomplishes two things: shows the ability to better maintain estimation consistency over most classic filtering approaches for single agents \cite{Wheeler2018RN}, and provides a convenient framework for a collaborative estimator.

This paper borrows from the above concepts for massive marginalization of a higher-rate, front-end filter that provides delta-pose inputs into a cooperative batch optimizer. However, in this case, there is also a desire to map the area the agents are traversing.  This requires the system to generate delta-pose inputs for the global batch optimization and maps that coincide with those measurements.  

To achieve this objective, the paper leverages and modifies versions of state of the art 3D SLAM algorithms \cite{6696650,Steinbrucker2011} which result in compact maplets and more compact delta-pose information for transforms between the traversed maplets. This collaborative navigation and mapping methodology provides a flexible approach that is scalable, communication efficient, and loss resilient for autonomous systems in indoor or underground settings. The key emphasis for this paper will be to minimize the amount of data that is required for sharing between collaborative agents while still incorporating a mapping capability in addition to the collaborative navigation.  

The rest of the paper is broken up as follows: Section \ref{sec:Overview} will provide a more detailed overview and introduce useful notation. Section \ref{sec:Methodology} will provide methodology and implementation details needed to implement the full system concept. Finally, Section \ref{sec:Results} will provide results followed by the conclusion in Section \ref{sec:conclusion}. 

\section{Overview \label{sec:Overview}}

Instead of sharing raw or nearly raw sensor data, or large scale maps themselves, this approach intentionally chooses to produce smaller, overlapping map elements, each with its own local reference frame/origin. In the case of a single agent, it can generate the transform, $T_{i,j \rightarrow i,j+1}$ between two consecutive and overlapping (both temporally and spatially) maplets, $M_{i,j}$ and $M_{i,j+1}$ (the redundant notation will be useful later), which consists of a highly marginalized odometry measurement in the form of a $3\times1$ delta-pose
and an associated $3\times3$ covariance for the 2-D ground vehicle case. This representation can easily be shared and leveraged in the second tier sparse batch optimization.

If cooperating agents are desired, then shared maplets with overlapping data can create a common reference frame. Let $i,k$ denote a pair of mapping agents that have shared and recognized overlapping maplets having indices $j,l$. We again denote the selected maplet data as $M_{i,j}$ and $M_{k,l}$ respectively and let $T_{i,j \rightarrow k,l}$ denote the unknown maplet-to-maplet transform to place vehicle $i$'s $j^{th}$ maplet origin (including orientation) in vehicle $k$'s $l^{th}$ maplet frame.

Further, it is worth keeping in mind that if range or bearing sensors are available between cooperating agents,
then tier I maplets can be coordinated. The maplet origins
have a range or bearing factor linking them,   
e.g. $\rho_{i,j,k,l}$ would be the range from vehicle $i$'s $j^{th}$
maplet to vehicle $k$'s $l^{th}$ maplet. This concept assumes the origin were assigned at the location of the agents within their current maplet at the time of the inter-agent measurement, which again, easily fits into the batch optimization scheme.

The dense RGBD and associated high-rate odometry, IMU, or other data has been fused into maplets as described in \cite{Tatavarti2017} with some additional modifications. Instead of operating on point clouds, this approach effectively looks for and builds the SLAM map out of planar, or near planar, sections of the point clouds. Additionally, instead of continuing to integrate and optimize the SLAM map with respect to an initial frame, the SLAM process is broken up to provide small, locally accurate maplets covering subsets of the navigable space. Those overlapping maplets, regardless of which vehicle produces them, can provide a host of maplet origin to maplet origin factors, similar to a delta-pose or odometry input, $T_{i,m \rightarrow k,n}$ and ideally an associated covariance or uncertainty factor, $P_{i,m \rightarrow k
,n}$, providing the necessary inputs to the sparse tier II global optimization. 

Notably, if regular inter-agent ranging is feasible, there is no requirement to share maplets at all in order to tie the global structure together.  However, without ranging or other factors, some maplet sharing will be required, but it can be approached judiciously to accommodate communication limitations. Maplets which present a high likelihood of overlapping with another agent's maplets, or maplets with particularly salient scene content can be prioritized. Finally, once an optimized global skeleton is made, and if desired, the maplets, either  own-agent or shared and at any fidelity desired (as long as the communication bandwidth can support), can be overlaid on the skeleton to generate the high resolution map of the space with global optimization of the denser map data itself. 

The intent here is to provide a framework that dramatically reduces the required communication and computational loads required to produce a functional product. Certainly if and when these resources are available, they could be leveraged, but it is not required. The approach also allows for graceful performance degradation (to single agent SLAM) when resources are limited while maintaining the ability to recover quickly if additional resources become available again. 

\section{\label{sec:Methodology}Methodology and Implementation}

The effort involves three main parts to create a multi-tier navigation approach. First, single agents will use dense odometry and point-cloud data to operate a modified 3D-SLAM algorithm. The dense 3D-SLAM information will be parameterized in an efficient manner and broken down into smaller, overlapping maplets which capture the geometry of the space with respect to a maplet origin. Next, maplets with overlapping coverage will be identified and used to provide the transform (and uncertainty) between their maplet frames. Finally, those transforms between maplet frames will be used in a pose- or factor-graph optimization to generate a skeleton for the global location and orientation of each maplet. 

\subsection{Generating Local Maplets}

Tier I of the system consists of mapping agents that build maps by processing dense RGBD sensor data. These agents use a customized version of the fast, multi-resolution, plane fitting algorithm in \cite{Tatavarti2017} to quickly compress sensed geometries focusing on extraction of large flat regions and extracting more detailed structures as a collection of quadrangular surface patches. Our enhancement to the algorithm described in \cite{Tatavarti2017} allows it to accept both time and space, i.e., memory, restrictions to generate an algorithm that is conformable to available computational resources.

The mapping agents build maps by aggregating sensed data over time into
keyframes.\footnote{Unlike the prior work already  mentioned, the keyframes discussed here occur at a much higher rate and there may be several keyframes within a given maplet. In the end, a single keyframe (or some other point) will be used as that maplet's origin which is more akin to the ``keyframe'' concept from \cite{Ellingson2020_Navigation}.}  Each keyframe includes a payload consisting of the aggregated plane, 3D point, and appearance (photometric) values extracted from sensed data since the creation of the previous keyframe. This generates a sparse collection of graph nodes having data from correlated, i.e., partially-overlapping, keyframes. Our method for keyframe creation follows the method described in Direct Visual Odometry (DVO-SLAM) \cite{6696650}. DVO-SLAM introduces new keyframes when a deterioration in the frame-to-keyframe tracking quality is detected, e.g., a large view change such as rotation or traversed distance. This is in contrast to \cite{6696650} which populates nodes with compute and resource hungry 3D point cloud data. We build compact maps consisting of quadrangular planar patches that adapt in size to achieve a specified goodness-of-fit criteria while simultaneously complying with specified computational (data size/bandwidth) constraints. 

\begin{figure*}[h]
\centering
\subfigure[]{
\begin{minipage}{.4\textwidth}
  \centering
    \fbox{\includegraphics[angle=0,origin=c,height= 0.9\linewidth, width=0.75\linewidth]{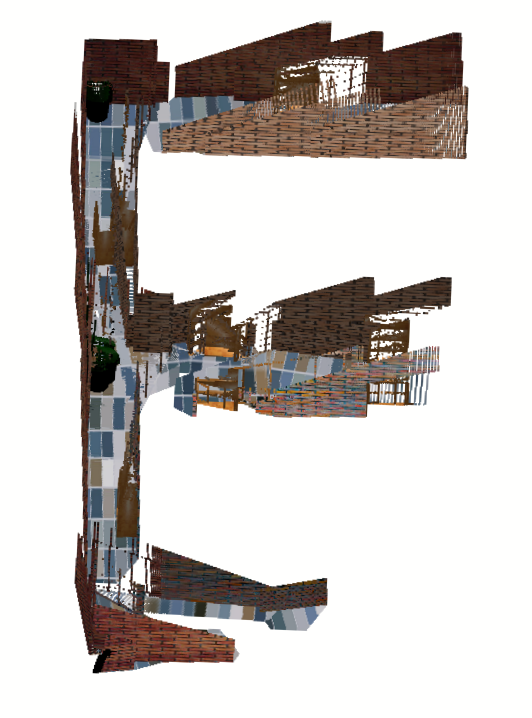}}
\end{minipage}
}
\hspace{.05\textwidth}
\subfigure[]{
\begin{minipage}{.4\textwidth}
  \centering
    \fbox{\includegraphics[angle=0,origin=c,height= 0.9\linewidth, width=0.75\linewidth]{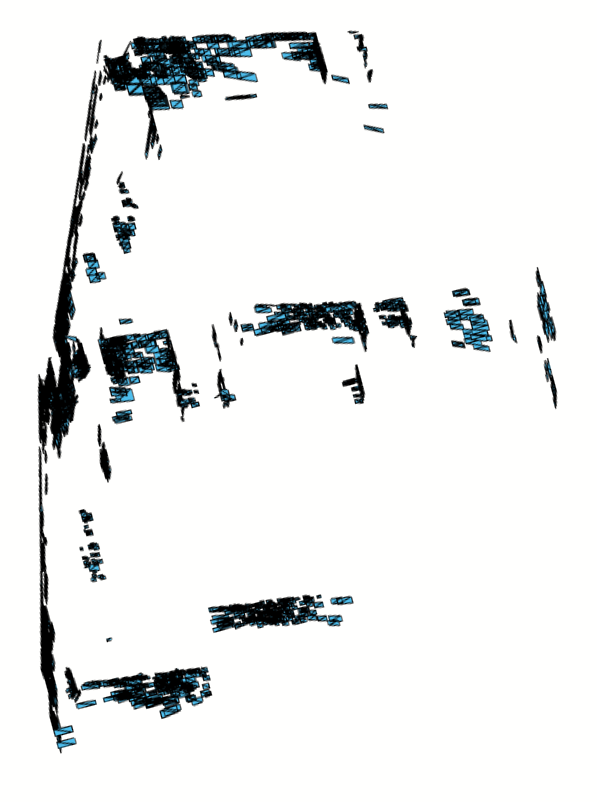}}
\end{minipage}
}
\caption{(a) High-density point cloud based DVO SLAM results. (b) Reduced order SLAM results with planar representations of surfaces in the scene using less than $1/100^{th}$ the data. \label{HD and planar}}
\end{figure*}

Under this formalization, graph nodes hold keyframe payloads consisting
of planar patches extracted from sensed data and graph edges correspond
to estimates of the delta-pose between keyframes having correlated node data. The generic algorithm for generating maplets is provided in Algorithm \ref{alg:graph-SLAM-to-Maplets}, and Figure \ref{HD and planar} shows both the high density point cloud as well as the marginalized planar representation of an example maplet which uses less than $1/100^{th}$ the data compared to the dense representation.  

It should also be noted that consideration for how and where to cut off maplets will significantly impact the overall system performance and should play a role in the implementation and execution of Algorithm \ref{alg:graph-SLAM-to-Maplets}. Some considerations are:
\begin{itemize}
    \item Because this implementation is relying on visual odometry and not wheel odometry (not used by DVO), the maplets can start to break down when the vehicle experiences large yaw. It would be prudent to limit the total yaw experience within a maplet (i.e. maintain its local accuracy and start a new, overlapping maplet as needed). \item Depending on the final use case, the global skeleton could be used to provide a rough map of the traversable space. In this case, breaking out maplets in a manner that allows the line from one maplet origin to the next maplet origin to pass through open space allows the skeleton to function as that low(est) fidelity map, so practical heuristics like setting maplet origins at intersections of hallways, etc. would be advised. 
    \item In order to avoid requiring extreme overlaps in maplets, setting maplets in a manner were there are salient features near the maplet boundary will improve the odds of identifying loop closures with maplets of other agents. 
\end{itemize}

\begin{algorithm}[h]
\SetAlgoLined
 \hspace{0.1cm}\textbf{Input:} graph-SLAM Map $m, G_m(N,E)$\\ 
 \hspace{0.1cm}\textbf{Output:} Map $m$, as maplets $M_{m,k}$\\
 \hspace{0.1cm}\textbf{Parameters:} $\kappa_{max}$ - curvature threshold\\
 $k=0$\;
 $N_s = G_m(N,E)$.origin();\hspace{1.0cm}// starting node\\
 \While{$G_m(N,E)$.hasNodes()}{
  $M_{m,k}.add(N_s)$; \hspace{0.5cm}// add start node,$N_s$, to $M_{m,k}$\\
  $G_m(N,E)$.remove($N_s$)\;
  $N_e = N_s\rightarrow$next;\hspace{1.0cm}// traverse edge, update end \\
 \While{  curvature(trajectory($N_s$,$N_e$)) < ${\kappa}_{max}$}{
  $M_{m,k}.add(N_e)$; \hspace{0.2cm}// add node $N_e$ to maplet $M_{m,k}$\\
  $G_m(N,E)$.remove($N_e$)\;
  $N_e = N_e\rightarrow$next;\hspace{0.5cm}// traverse edge, update end\\
 }
 $N_s = N_e$; 
 $k = k + 1$;\hspace{1cm}// start new maplet $M_{m,k+1}$\\
 }
 return $M_{m,1,...,k-1}$\;
 \caption{\label{alg:graph-SLAM-to-Maplets}graph-SLAM to maplet-SLAM; A Trajectory Curvature Decomposition}
\end{algorithm}

\subsection{Maplet to Maplet Transforms}\label{subsec:maplet-pair-alignment}

Our approach for creating and aligning maplet pairs is accomplished by two algorithms: one that creates maplet pairwise match hypotheses (Algorithm \ref{alg:maplet-pair-alignment}) and one that evaluates match hypotheses to align maplet pairs (Algorithm \ref{alg:iterative-closest-algebraic-plane}). 
The maplet pairwise hypothesis algorithm postulates pairwise matches between maplets. In this article, hypotheses are created at un-mapped maplet frontiers where additional unknown map geometry is likely to exist. A 2D polygon Region Of Interest (ROI) is used to select candidate planes to match with similar regions from other maplets. %Maplets and the alignment between two maplets are shown in Fig. \ref{2maplets}.

Each pairwise maplet hypothesis is evaluated using the Iterative Closest Algebraic Plane (ICaP) algorithm (Algorithm \ref{alg:iterative-closest-algebraic-plane}). This algorithm uses the compact representation of the world in terms of planar algebraic surfaces, i.e., surfaces having equation $ax+by+cz+d = 0$, to establish the likelihood that a given hypothesized maplet pair can be aligned with the intent to piece together large geometric map regions in a manner similar to puzzle-solving. This is accomplished by minimizing an error functional that solves for both the correspondence of planar surfaces between the maplets and the Euclidean transform that aligns these algebraic surfaces. The magnitude of the algebraic alignment error then serves as a goodness-of-fit metric to validate or refute the maplet pair hypotheses. 

While the descriptions of Algorithms \ref{alg:maplet-pair-alignment} and \ref{alg:iterative-closest-algebraic-plane} show program flow, our mathematical solution for alignment of corresponding algebraic plane equations is described in the following sections. These steps are performed inside the \emph{alignPlanePairs()} function from Algorithm \ref{alg:iterative-closest-algebraic-plane} which includes the following statement:
\begin{equation}
\mathbf{T}_{i+1} = alignPlanePairs(matchIndexPairs,\pi_j,\pi_l);
\end{equation}

This function computes the optimal alignment (in the least squares sense) between planes that are hypothesized to have the same equation up to an unknown Euclidean transformation. For our derivation we denote $\pi_j$ and $\pi_l$ as two collections of $N$ plane equations. Equation (\ref{eq:lsq_plane_alignment}) expresses the optimization problem at hand. Here we seek to estimate the transformation $\widehat{\mathbf{T}}_{i \rightarrow j}$ that takes the planes of $\pi_l$ into the coordinate system of planes $\pi_j$. Note that Euclidean transformations, when applied to planes, follow the transformation rule $\pi^{'} = (\mathbf{T}^{-1})^t\pi$ for $\pi^t=\begin{bmatrix}a & b & c & d]\end{bmatrix}$.
\begin{equation}
\label{eq:lsq_plane_alignment}
    \widehat{\mathbf{T}}_{i \rightarrow j} = \min_{\mathbf{T}_{i \rightarrow j}}\sum_{\{i,j\} pairs}	\|\pi_j-(\mathbf{T}_{i,j}^{-1})^t\pi_i\|^2 
\end{equation}

As in least squares point alignment \cite{Umeyama91}, the solution is obtained by decomposing the problem into two steps: solve for the rotation,  $\widehat{\mathbf{R}}$, and solve for the translation, $\widehat{\mathbf{p}}$. After solving for these variables separately, the optimal transformation can be constructed as shown in Equation (\ref{eq:T_internals}).
\begin{equation}
\label{eq:T_internals}
\widehat{\mathbf{T}} = \begin{bmatrix}
            \widehat{\mathbf{R}} & \widehat{\mathbf{p}}\\ 
            \mathbf{0} & 1\\
        \end{bmatrix}
\end{equation}

We solve for the rotation, $\widehat{\mathbf{R}}$, that aligns the orientations of corresponding planes by finding the rotation that maximally aligns their normals, i.e., the vectors formed by the first three coefficients of each matching plane pair. To do so, we form a covariance matrix of the matching plane normal vectors as shown in Equation (\ref{eq:normal-covariance}).
\begin{equation}
\label{eq:normal-covariance}
C_{\mathbf{n}} = \sum_{\{i,j\} pairs} \mathbf{n}_i\mathbf{n}_j^t 
\end{equation}

We then decompose the covariance matrix using singular value decomposition, $svd()$, to compute the orthogonal transformation that minimizes the squared error, aligning corresponding plane normals as shown in Equation (\ref{eq:svd-decompostion}).
\begin{equation}
\label{eq:svd-decompostion}
\mathbf{U}\Lambda\mathbf{V}^t = svd(C_{\mathbf{n}})
\end{equation}

As pointed out in related literature on point alignment \cite{Umeyama91}, the optimal rotation from Equation (\ref{eq:svd-decompostion}) could require a reflection. To find the optimal right-handed rotation, we must replace the smallest eigenvalue with $det(\mathbf{U}\mathbf{V}^t)$ to obtain the right handed rotation incurring smallest additional error beyond our original solution as shown in Equation (\ref{eq:rotation-reflection}).
\begin{equation}
\label{eq:rotation-reflection}
\widehat{\mathbf{R}} = \mathbf{U}\begin{bmatrix}
            1 & 0 & 0\\ 
            0 & 1 & 0\\ 
            0 & 0 & det(\mathbf{U}\mathbf{V}^t)\\
        \end{bmatrix}\mathbf{V}^t 
\end{equation}

Using our solution for the optimal rotation, we then solve for the translation component of the transform that brings the two plane sets into final alignment. To do so, we form a matrix of the aligned normal vectors of the set $\pi_j$, $\widehat{\mathbf{N}}_j$. We also form vectors of the fourth coefficients of the corresponding planes to be aligned; $\mathbf{d_l}$ and $\mathbf{d_j}$. The mathematical representation for these variables are shown in the Equations (\ref{eg:vectors-for-translation-solution}).
\begin{equation}
\begin{matrix}
\label{eg:vectors-for-translation-solution}
\widehat{\mathbf{N}}_i^t = \widehat{\mathbf{R}}\begin{bmatrix}
            \mathbf{n}_{i_0} & \mathbf{n}_{i_1} & ... & \mathbf{n}_{i_N}
        \end{bmatrix}\\
        \mathbf{d_j}^t = 
        \begin{bmatrix}d_{j_0} & d_{j_1} & ... & d_{j_N} \end{bmatrix}\\
        \mathbf{d_i}^t = 
        \begin{bmatrix}d_{i_0} & d_{i_1} & ... & d_{i_N} \end{bmatrix}\\
        \end{matrix}
\end{equation}

Using these matrices and vectors we can write the explicit least squares solution for the translation component, $\widehat{\mathbf{p}}$, as shown in Equation (\ref{eq:solution-for-translation}).
\begin{equation}
\label{eq:solution-for-translation}
\widehat{\mathbf{p}} = ({\widehat{\mathbf{N}}_i^t}\widehat{\mathbf{N}}_i)^{-1}{\widehat{\mathbf{N}}_i}^t(\mathbf{d_i} - \mathbf{d_j})
\end{equation}

\begin{algorithm}[h]
\SetAlgoLined
\hspace{0.1cm}\textbf{Input:} Maplets $M_{i,j},M_{k,l}$ and polygon ROI, $P$,\\ 
\hspace{0.1cm}\textbf{Output:} $\mathbf{T}_{i,j \rightarrow k,l}$ the transform aligning $M_{i,j}$ to $M_{k,l}$ \\
 $\pi_j,\pi_l = \{\emptyset\}$; \hspace{0.2cm}// sets of candidate planes to match\\
\ForAll{$\pi \in M_{i,j}$.getPlanes()} {
 \If{$P$.contains($\pi$)}{
 $\pi_j$.add($\pi$)\;
 }
 }
 \ForAll{$\pi \in M_{k,l}$.getPlanes()} {
   \If{P.contains($\pi$)} {
   $\pi_l$.add($\pi$)\;
   }
 }
\If{size$(\pi_j) < 3$ or size$(\pi_l) < 3$}{
  return $null$;\hspace{1cm}// solution not unique\\ 
 }
\eIf{size$(\pi_j) < size(\pi_l)$}{
$\mathbf{T}_{k,l \rightarrow i,j} =$ iterativeClosestPlane($\pi_j,\pi_l,\mathbf{I}$)\;
return  $\mathbf{T}_{k,l \rightarrow i,j}^{-1}$\;
}{
$\mathbf{T}_{i,j \rightarrow k,l} =$ iterativeClosestPlane($\pi_l,\pi_j,\mathbf{I}$)\;
return  $\mathbf{T}_{i,j \rightarrow k,l}$\;
}
 \caption{\label{alg:maplet-pair-alignment}Maplet-Pair Alignment}
\end{algorithm}

\begin{algorithm}[h]
\SetAlgoLined
\hspace{0.1cm}\textbf{Input:} Two sets of plane coefficients $\pi_j,\pi_l$ and initial transform $\mathbf{T}_0$\\
\textbf{Assume:} size($\pi_j$) > 3, size($\pi_l$) > 3, size($\pi_j$) $\geq$ size($\pi_l$)\\
\hspace{0.1cm}\textbf{Output:} $\widehat{\mathbf{T}}$ the transform minimizing Equation (\ref{eq:lsq_plane_alignment})\\
\hspace{0.1cm}\textbf{Parameters:} $\tau_T$ - threshold for convergence\\
$\mathbf{T}_i = \mathbf{T}_0$\;
$\Delta \mathbf{T} = \infty$\;
N = size($\pi_l$);\hspace{1cm} // number of matches to compute\\
\While{$\Delta \mathbf{T} \geq \tau_T$}{
 $matchIndexPairs = \{\emptyset\}$ \hspace{0.1cm}// closest plane index-pairs\\
 \ForAll{${l_0} \in 1,2,...,N$} {
 $j_0 = \min_{j}\|\pi_{j}-(\mathbf{T}_{i}^{-1})^t\pi_{l_0}\|^2$; \\
 matchIndexPairs.add($\{l_{0},j_{0}\}$)\;
 } 
 // solve Equation (\ref{eq:lsq_plane_alignment}) as described in article \S \ref{subsec:maplet-pair-alignment}\\
 $\mathbf{T}_{i+1} = $alignPlanePairs(matchIndexPairs,$\pi_j,\pi_l$);\\
 $\Delta \mathbf{T} = \|\mathbf{T}_{i+1}-\mathbf{T}_{i}\|^2$;\\
 $\mathbf{T}_{i} = \mathbf{T}_{i+1}$\;
 }
return  $\mathbf{T}_{i}$\;
 \caption{\label{alg:iterative-closest-algebraic-plane}Iterative Closest Algebraic Plane (ICaP)}
\end{algorithm}

\textbf{Practical considerations:} There are several important criteria necessary for the algebraic alignment solution to work well in practice: 
\begin{itemize}
\item We recommend the following unique representation for planes: ensure all measured surfaces have normals with a negative Z component in the coordinate system of the sensor and ensure the plane coefficients are in Hesse normal form, i.e., the first three coefficients of the plane equation $ax+by+cz+d=0$ have unit length. 
Under these circumstances, the resulting implicit plane equations are unique, have beneficial computation qualities, and their first three coefficients carry geometric information as the outward pointing unit length normal to the plane surface.

\item For $\mathbf{T}_{i,j \rightarrow k,l}$ to be leveraged properly as a factor or edge for either a factor- or pose-graph, we need an appropriate uncertainty for the transform. In general, we can assume all the transforms have about the same uncertainty, thus a generic covariance can be applied. While this will likely work in a lot of cases, the transform's accuracy is highly dependent on the quality of the maplets and their geometries. An approach similar to \cite{willis2016} or \cite{Anderson2019} could be used with the ICaP algorithm to assess the relative quality/accuracy of the transpose.  

\item Finally, maplets could be modified to include a broader range of geometric parameterizations to allow for curvature or other common feature types. Further, for specifically salient RGB data, appearance as texture for the surfaces could drastically improve alignment and/or reduce the necessary number of matched structures. 
\end{itemize}

\begin{figure*}[htbp]
\centering
\subfigure[]{
\begin{minipage}[t]{.45\textwidth}
  \centering
    \fbox{\includegraphics[angle=0,origin=c,height=.6\linewidth, width=0.9\linewidth]{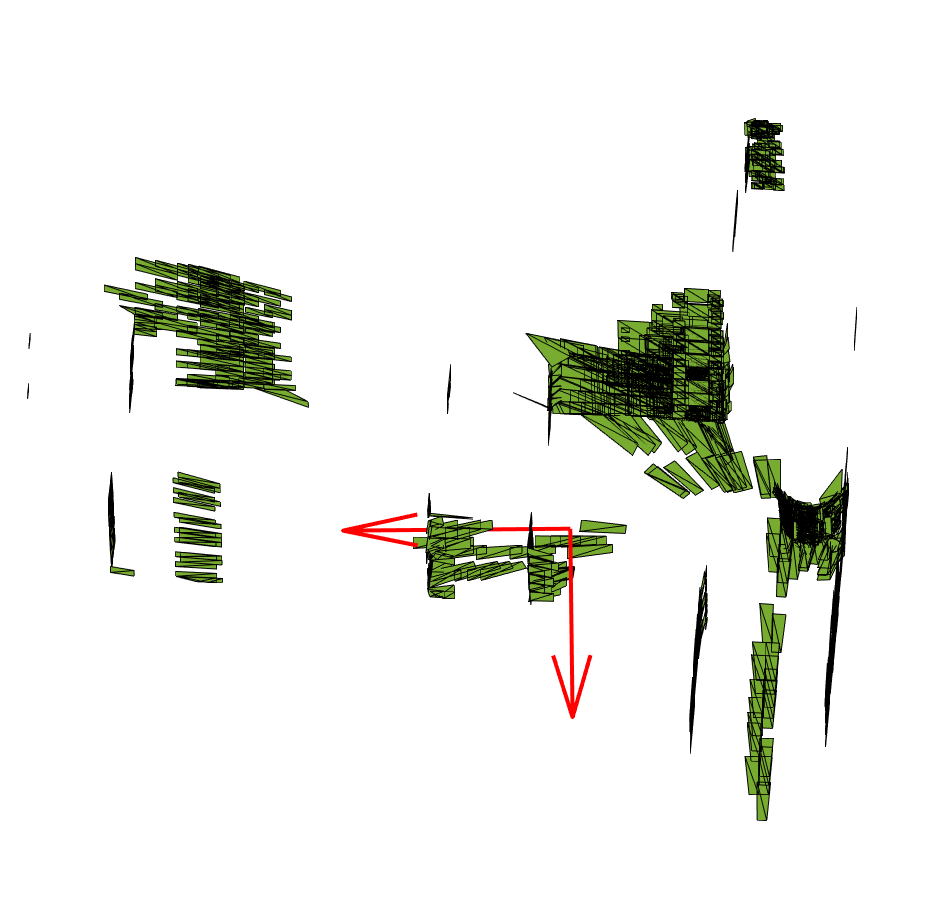}}
\end{minipage}
}
\subfigure[]{
\hspace{.001\textwidth}
\begin{minipage}[t]{.45\textwidth}
  \centering
    \fbox{\includegraphics[angle=0,origin=c,height=.6\linewidth, width=0.9\linewidth]{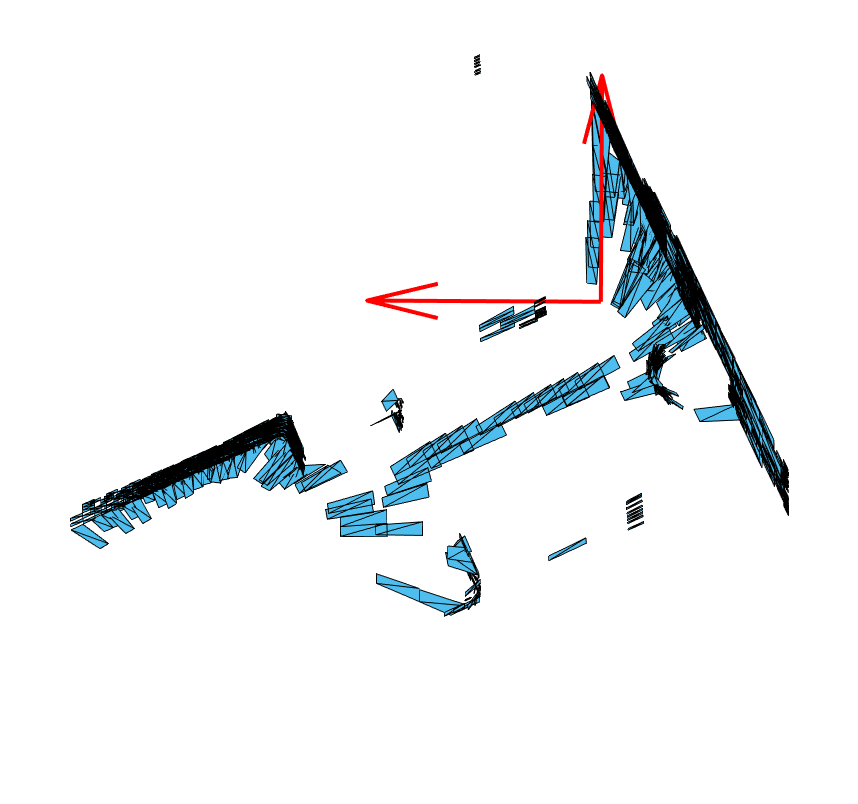}}
\end{minipage}
}
\\
\subfigure[]{
\begin{minipage}[t]{.45\textwidth}
  \centering
    \fbox{\includegraphics[angle=0,origin=c,height=.6\linewidth, width=0.9\linewidth]{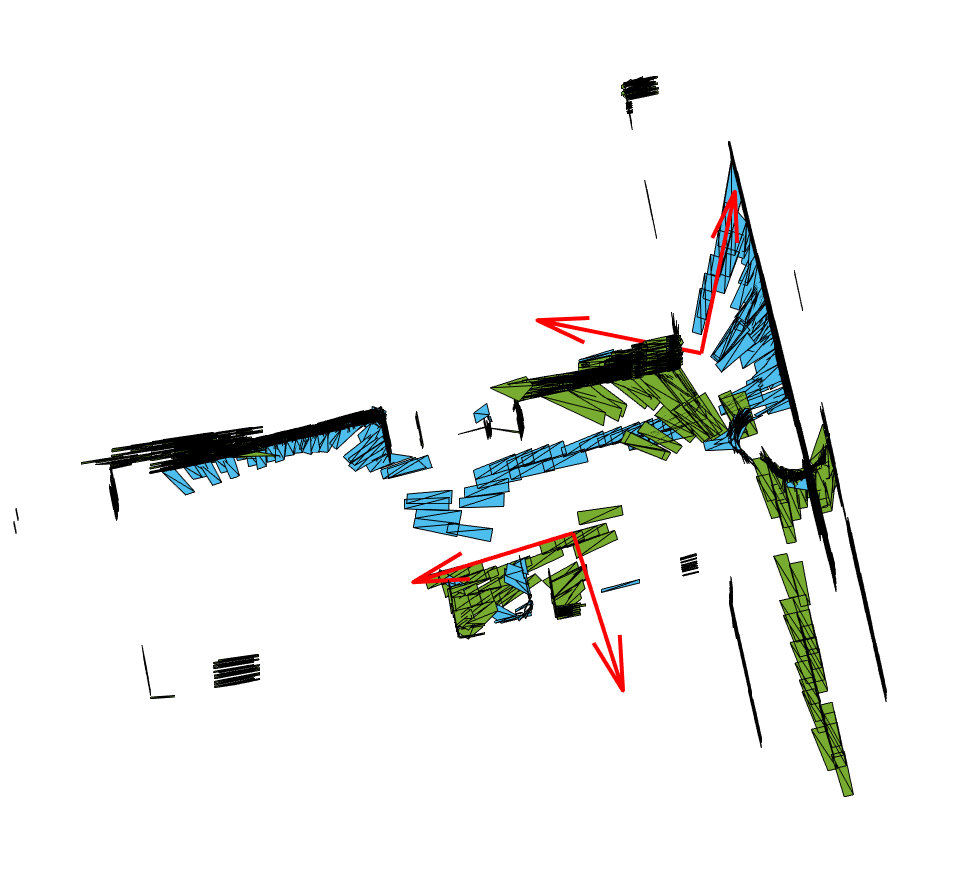}}
\end{minipage}
}
\subfigure[]{
\hspace{.001\textwidth}
\begin{minipage}[t]{.45\textwidth}
  \centering
    \fbox{\includegraphics[angle=0,origin=c,height=.6\linewidth, width=0.9\linewidth]{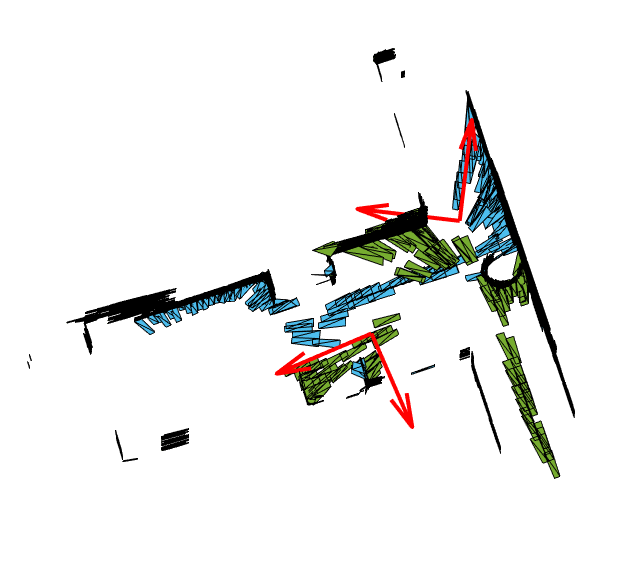}}
\end{minipage}
}
\caption{Example maplets and alignment: (a) Maplet 1 and (b) Maplet 2, which have overlapping data. (c) coarse alignment of maplets and (d) refined alignment after ICaP, which provides the necessary delta-pose, or transform, $T$, from Maplet 1 to Maplet 2. \label{2maplets}}
\end{figure*}

Figure \ref{2maplets} shows two unique maplets generated as described in Algorithm \ref{alg:graph-SLAM-to-Maplets}. In this case they overlap and the figure depicts a rough alignment followed by a refined alignment using Algorithm \ref{alg:maplet-pair-alignment} and Algorithm \ref{alg:iterative-closest-algebraic-plane}.

\subsection{Optimizing the Map Skeleton}
Generally, vehicle wheel odometry is an accurate measurement source for the exploration of both small and large areas.  However, maps generated from odometry often exhibit drift after traversing long hallways or after the exploration of very large structures.  In this section, we present the tier II batch optimization routine that operates on the maplet to maplet transforms mentioned earlier. The highly marginalized delta-poses are used to create a pose-graph of the links between maplets, effectively a skeleton representation of the cooperatively explored space. 

If maplet $M_{i,j}$ has an estimated pose $X_{i,j}$, within the broader optimization, and if it is related to maplet $M_{k,l}$ by the transform $T_{i,j \rightarrow k,l}$, then the pose $X_{k,l}$ can be generated. These pose estimates, $X$, and inter-pose, or delta-pose measurements, $T$, and their associated covariances, $P$, provide the necessary inputs.  

The batch optimization operates on two sets of data: the relative poses between maplet pairs and any loop closures that may have been determined, which are a highly studied problem space in batch optimization. Occasional loop closures, generated from a vehicle recognizing that it returned to a location previously explored by itself or another vehicle, are often sufficient to correct error in the vehicle odometry measurements.  In the case of a very simple graph, e.g. single vehicle with no loop closures, then the optimizer best guess of $X_{k,l}$, or any other pose, is simply a chained sequence of transforms stacking each subsequent pose into the prior pose frame reaching back to the nominal global/referenced frame. However, as soon as loop closures (either back to a vehicle's prior pose or between multiple vehicles) are introduced, the poses are shifted to best minimize the probability-weighted errors within the competing constraints.

For the results presented later in the paper, we selected the PoseSLAM batch algorithm which is part of the GTSAM implementation \cite{Dellaert12}.  Alternatively, another commonly used batch optimization tool was proposed by \cite{Kummerle2011} which was used to provide improvements for batch robustness when operating on delta-pose data.  For cases of poor initial conditions, i.e. when cooperative agents are not sure of their initial global or relative poses, the work by \cite{Jackson2019Edge} may provide an attractive solution.

Because a publicly-available batch implementation was used in this paper  the algorithm description is omitted \cite{Dellaert12}. Instead, there are some key considerations that should be discussed for how to best coordinate between agents in order to provide the batch algorithm of choice with the most beneficial inputs while maintaining the low-bandwidth assumptions on communications:
\begin{itemize}
    \item When cooperative agents come into communication range, they should both share a list of the most recent delta-pose measurements they have from all other agents, then share any data from that list which the other agent is missing. 
    \item Second, they should share the list of recent loop closures, generally transforms between maplets generated by more than one agent. 
    \item Both lists are simply a set of extremely compact indexes and after the missing transforms and associated covariances are shared, both agents will have matching global/skeleton optimizations. 
    \item Maplets that are deemed to have particularly salient features, or maplets that, based on the optimized skeleton, appear to be likely candidates of a loop closure would be shared next, communications permitting. 
    \item Finally, additional maplets can be shared when bandwidth is available. There is no future computation cost required after sharing the maplets (they can be overlaid on the skeleton, but are not required to be operated on) and the more maplets (at 15KB a piece) a vehicle has, the more possible avenues for future loop closures exist.
\end{itemize}

It should be noted that this particular approach clearly allows for extended communications dropouts, or even complete loss of some cooperative agents. The key is that the most important information, the delta-poses, are extremely bandwidth-friendly and can be shared rapidly to ``catch up'' and then denser data, i.e. maplets, can be shared selectively if bandwidth allows.

As mentioned earlier, the global skeleton can provide a rough map of the traversable space.  Consider the case of two vehicles exploring a maze-like structure and one of the vehicles locating the exit.  Knowing the global position of the exit may not be helpful to the second vehicle if it cannot determine a navigable path, and sharing a global map of the structure, either through a global map or maplets, may consume significant communication resources.  Sharing the skeleton gives the second vehicle a rough path to follow, first through the area that it has explored to a loop closure location seen by both vehicles and then through the area that the second vehicle has explored to the exit.  This approach is clearly applicable to multiple vehicles while remaining communication limited.

\section{Results \label{sec:Results}}

\begin{figure*}[htp]
\centering
\subfigure[]{
\begin{minipage}{.29\textwidth}
  \centering
    \fbox{\includegraphics[angle=0,origin=c,height = 1.2\linewidth, width=1.07\linewidth]{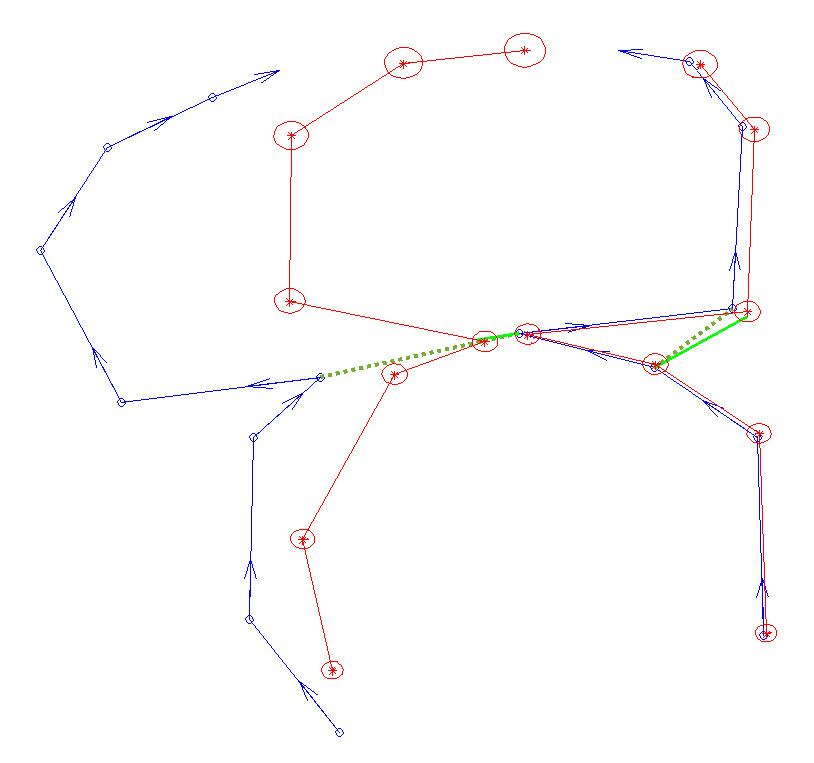}}
\end{minipage}
}
\hspace{.02\textwidth}
\subfigure[]{
\begin{minipage}{.29\textwidth}
  \centering
    \fbox{\includegraphics[angle=0,origin=c,height = 1.2\linewidth, width=1.07\linewidth]{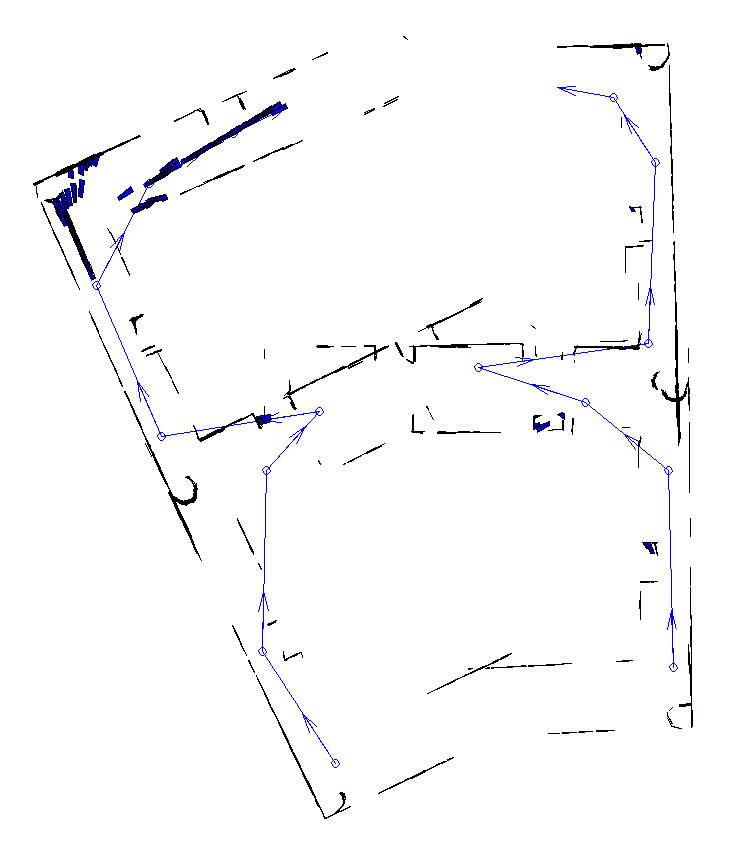}}
\end{minipage}
}
\hspace{.02\textwidth}
\subfigure[]{
\begin{minipage}{.29\textwidth}
  \centering
    \fbox{\includegraphics[angle=0,origin=c,height = 1.2\linewidth, width=0.9\linewidth]{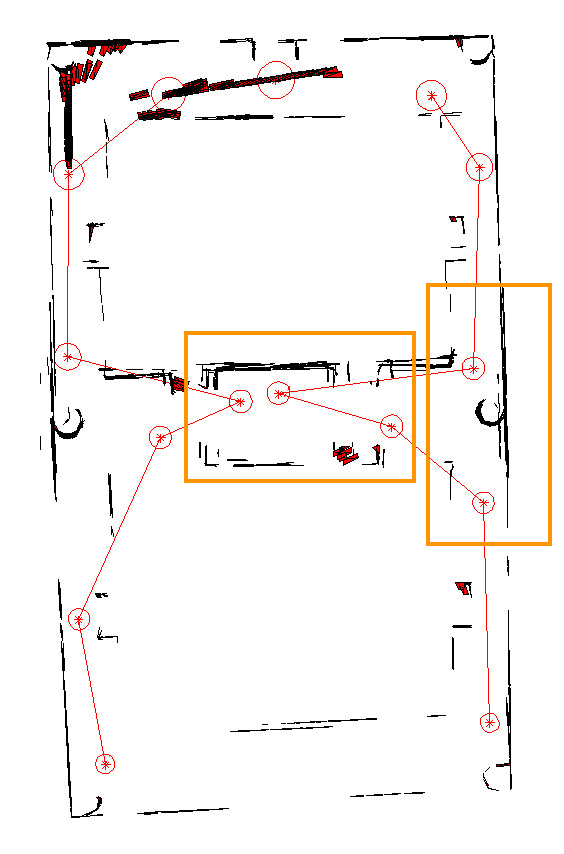}}
\end{minipage}
}
%\subfigure[]{
%\begin{minipage}{.3\textwidth}
%  \centering
%    \fbox{\includegraphics[angle=0,origin=c,height = 1.2\linewidth, width=0.9\linewidth]{img2/optimized_map.png}}%
%\end{minipage}
%}
\caption{(a) Unoptimized (blue) and optimized (red) map skeleton. Green dashed and green solid lines indicate the maplet-based loop closures pre- and post-optimization, respectively. (b) Maplet-overlays of independent vehicle delta-pose map skeletons. (c) Map generated after loop closure (green lines) included in skeleton optimization, the orange rectangular areas show the maplets alignment from optimization.\label{maps}}
\end{figure*}

The proposed system was exercised in simulation with an appropriate RGBD sensor model and associated vehicle trajectory through an indoor-like environment. The majority of the results have already been shown while illustrating the concepts and methodologies in Section \ref{sec:Methodology}. Here the unoptimized and optimized skeleton maps are shown in Figure \ref{maps}a while the same results with their associated maplet overlays are shown in Figure \ref{maps}b and \ref{maps}c. 

%\begin{figure}[h]
%\centering
%\hspace{.001\textwidth}
%\begin{minipage}{.5\textwidth}
%  \centering
%    \fbox{\includegraphics[angle=0,origin=c,width=0.7\linewidth]{img2/optimized_back-end.png}}
%\end{minipage}
%\caption{Unoptimized (blue) and optimized (red) map skeleton. Green dashed and green solid lines indicate the maplet-based loop closures pre- and post-optimization, respectively. }
%\label{result1} 
%\end{figure}

Table \ref{tab:message size} shows the bandwidth required for the data streams of the planar and point cloud optimization implementations. The naive implementations would require approximately 16KB per maplet, or 2MB of data for the dense point cloud maps, and while some saving could be found, it is clear the planar representation is significantly more compact.

Further, the skeleton was generated using 15 transforms and associated covariance estimates (from the maplet to maplet alignment scheme presented earlier) and those delta-pose messages consisted of doubles in a $3\times1$ and $3\times3$ format, plus indexing information, for a total of approximately 0.1KB of data per delta-pose in the optimization. Again, the map loop closures in this case were generated by the identification of overlapping scenes in shared maplets, other concepts, including inter-vehicle ranging would not require shared maplets unless specifically desired.  Note that in Figure \ref{maps} only a single loop closure between agents was used, but because the transform has both translation and rotation data, it was still able to align the full map. Even a limited number of loop closures or range measurements can provide strong ties between the vehicles. If two vehicles never communicate directly, it is quiet feasible for them to be connected though a mutual neighbor (or chain of neighbors) and the delta-pose inputs they share providing almost identical back-end graph structures and a shared understanding of their relative poses.  

The results shown here, using delta-pose and with every maplet shared, requires about 130 times less data to be transmitted than the raw point cloud representation otherwise would. When one considers the additional cooperating agents, the fact that the individual vehicle odometry is highly accurate and thus loop closures would not be required particularly frequently, and that communications may be highly intermittent, the ability to operate on the delta pose measurements alone becomes increasingly appealing.

% \begin{figure*}[h]
% \centering
% \hspace{.001\textwidth}
% \begin{minipage}{.5\textwidth}
%   \centering
%     \fbox{\includegraphics[angle=0,origin=c,width=\linewidth]{img/withOverlay.jpg}}
% \end{minipage}
% \caption{Global optimization with maplet overlay. \kb{To be updated soon with results from GTSAM and SLAM overlay}. \label{result1}}
% \end{figure*}

\begin{table}[h]
\centering
\begin{tabular}{llccc}
\hline 
 Vehicle& &min&max&mean\\
\hline
\multirow{2}*{Left}&pointclouds&2.15MB&2.19MB&2.17MB\\
%\cline{2-6}
 &maplets&2.38KB&36.64KB&16.28KB\\
\hline 
\hline
\multirow{2}*{Right}&pointclouds&2.15MB&2.18MB&2.16MB\\
%\cline{2-6}
 &maplets&3.48KB&33.39KB&14.89KB\\
\hline
\end{tabular}
\vspace{1.5mm} %
\caption{Bandwidth used by SLAM algorithms showing significant reduction using planar representation for maplet features compared to point cloud representations. }
\label{tab:message size}
\end{table}

\section{Conclusion \label{sec:conclusion}}

This article introduced an approach to facilitate cooperative exploration and mapping of large-scale, near-ground, underground, or indoor spaces. The concept was to break up the mapping process from a large, monolithic effort where cooperating agents are required to share large quantities of sensor data, to a tiered approach were only modest amounts of data must be shared. 

A unique modification of RGBD SLAM was detailed, where the SLAM was done using compact planar representations of the space. The SLAM maps were further broken down into small, overlapping maplets, each with its own unique reference frame/origin. Additionally, a ``closest plane'' method was introduced to generate maplet to maplet transforms that can be used in a classic batch pose optimization problem. Finally, the maplets can be overlaid on the batch results to generate a global map of the space.  

The concept and associated algorithms were shared in detail and exercised on simulated RGBD data in an indoor-like environment and shown to produce a representative map of the space without assuming an uninterrupted or high bandwidth communication capability.  

\bibliographystyle{ieeetr}
\bibliography{2020_ION_PLANS_MapletSLAM}

%\section{EXTRA}

%n order to perform batch optimization, the relative pose between hypothesized corresponding maplet pairs needs to be determined.  Let $\overbar{\mathbf{T}}_i$ denote the pose of maplet $\mathbf{M}_i$ in a map. Let $\overbar{\mathbf{T}}_j$ denote the pose of maplet $\mathbf{M}_j$ in a map. For batch optimization, we extract the relative pose of the maplet pair $\{i,j\}$ by 
%\begin{equation}
%    \overbar{\mathbf{T}}_j = \overbar{\mathbf{T}}_{i \rightarrow j}\overbar{\mathbf{T}}_{i}  
%\end{equation}
%\begin{equation}
%\label{eq:relative_pose}
%    \overbar{\mathbf{T}}_{i \rightarrow j}  = \overbar{\mathbf{T}}_j\overbar{\mathbf{T}}_i^{-1}
%\end{equation}
%Note that in Equation (\ref{eq:T_internals}), $\mathbf{T}$ was defined as a matrix containing both a rotation and a translation.  Therefore, to obtain the relative pose $\overbar{\mathbf{T}}_{i \rightarrow j}$ the inverse rotation must be taken before determining the relative translation.

\end{document}